\documentclass[10pt,twocolumn,letterpaper]{article}

\usepackage{iccv}              

\usepackage{latexsym}
\usepackage{amssymb}
\usepackage{amsmath}
\usepackage{amsthm}
\usepackage{booktabs}
\usepackage{enumitem}
\usepackage{graphicx}
\usepackage{color}
\usepackage{xcolor}

\DeclareUnicodeCharacter{2009}{\,}
\usepackage{tabularx}
\usepackage[table]{xcolor}
\usepackage{amssymb}
\usepackage{pifont}
\newcommand{\xmark}{\ding{55}} 

\definecolor{lightblue}{HTML}{A1E3F9}   
\definecolor{lightgreen}{HTML}{CAE0BC}   

\definecolor{mildgreen}{HTML}{5D9B63}

\definecolor{lrline}{HTML}{F9D3D3}   
\definecolor{lyheader}{HTML}{FFF6CC} 

\usepackage{booktabs, multirow, xcolor, pifont, siunitx}
\usepackage[table]{xcolor}

\definecolor{iccvblue}{rgb}{0.21,0.49,0.74}
\usepackage[pagebackref,breaklinks,colorlinks,allcolors=iccvblue]{hyperref}

\def\paperID{54} 
\def\confName{ICCV}
\def\confYear{2025}

\title{Not Only Grey Matter: OmniBrain for Robust
Multimodal Classification of Alzheimer's Disease}

\author{Ahmed Sharshar\thanks{Corresponding author. Email: ahmed.sharshar@mbzuai.ac.ae} \and  Yasser Ashraf \and Tameem Bakr \and Salma Hassan \and Hosam Elgendy \and Mohammad Yaqub \and Mohsen Guizani \and
Mohamed bin Zayed University of Artificial Intelligence (MBZUAI), Abu Dhabi UAE.
\\
\{firstname.lastname\}@mbzuai.ac.ae}

\begin{document}
\maketitle
\begin{abstract}
Alzheimer's disease affects over 55 million people worldwide and is projected to more than double by 2050, necessitating rapid, accurate, and scalable diagnostics. However, existing approaches are limited because they cannot achieve clinically acceptable accuracy, generalization across datasets, robustness to missing modalities, and explainability all at the same time. This inability to satisfy all these requirements simultaneously undermines their reliability in clinical settings. We propose \textit{OmniBrain}, a multimodal framework that integrates brain MRI, radiomics, gene expression, and clinical data using a unified model with cross-attention and modality dropout. \textit{OmniBrain} achieves $92.2 \pm 2.4\%$ accuracy on the ANMerge dataset and generalizes to the MRI-only ADNI dataset with $70.4 \pm 2.7\%$ accuracy, outperforming unimodal and prior multimodal approaches.  Explainability analyses highlight neuropathologically relevant brain regions and genes, enhancing clinical trust. OmniBrain offers a robust, interpretable, and practical solution for real-world Alzheimer’s diagnosis.
\
\end{abstract}    
\section{Introduction}
\label{sec:intro}

Alzheimer’s disease (AD) is a chronic neurodegenerative disorder that progressively erodes memory, reasoning, and communication. It accounts for 60-70\% of dementia cases worldwide \cite{who_dementia_2023}, affecting about 55 million people in 2020 and expected to top 139 million by 2050 \cite{who_dementia_2023}. In the United States, nearly 7 million people live with AD, a figure projected to almost double by mid‑century \cite{alz_facts_2024}. Dementia already costs more than \$1 trillion annually and could reach \$2 trillion by 2030, underscoring the need for accurate and efficient diagnostics.

AD evolves from mild cognitive impairment (MCI), subtle cognitive deficits, to severe dementia; symptom‑free individuals are labeled cognitively normal controls (CTL). Distinguishing CTL, MCI, and AD is difficult because symptoms overlap, onset is gradual, and definitive biomarkers remain scarce. Many deep‑learning models train on homogeneous cohorts and therefore fail to generalise \citep{aghdam_survey_2025}.

Despite rapid advances in medical AI, most Alzheimer’s diagnostic systems remain limited in practical scope. As summarized in Table \ref{tab:methodology_comparison}, prior studies typically focus on a single data modality (MRI scans), neglecting key biomarkers (e.g., gene expression, cognitive scores) crucial for reliable AD classification. Even recent multimodal models rarely address two essential challenges: (1) the frequent absence of clinical or genetic data in real-world settings, and (2) the need for robust validation across multiple datasets to ensure generalizability. Furthermore, explainability—a prerequisite for clinical adoption—remains underexplored. These gaps have collectively hindered translation from research to routine care, motivating the development of a comprehensive framework that is not only multimodal, but also resilient to missing data, validated across cohorts, and interpretable by design.

The proposed framework addresses these limitations by introducing a comprehensive, multimodal framework that fuses structural MRI information with radiomics, meta information (e.g., APOE), cognitive scores (e.g., MMSE), and gene expression features to classify CTL, MCI, and AD robustly while being able to perform if one of these modalities is missing constantly. Specifically, we leverage:
\begin{itemize}[leftmargin=*]
    \item \textit{Foundation models} (AnatCL, y-Aware InfoNCE) for anatomical MRI representations and an \textit{FT-Transformer} for tabular data (radiomics, metadata, gene expression), integrating these via an attention-based fusion mechanism.
    \item A \textit{missing-modality approach} that incorporates genes and clinical metadata during training but allows for inference with or without these inputs, reducing real-world data constraints.
    \item \textit{Cross-dataset validation} on ADNI—showcasing state-of-the-art results on ANMerge and strong generalization to different data distributions. We also employ interpretable analytics (e.g., radiomics) to highlight key features and enhance clinical trust.
\end{itemize}

\begin{figure*}[t!]
    \centering
    \includegraphics[width=0.75\textwidth]{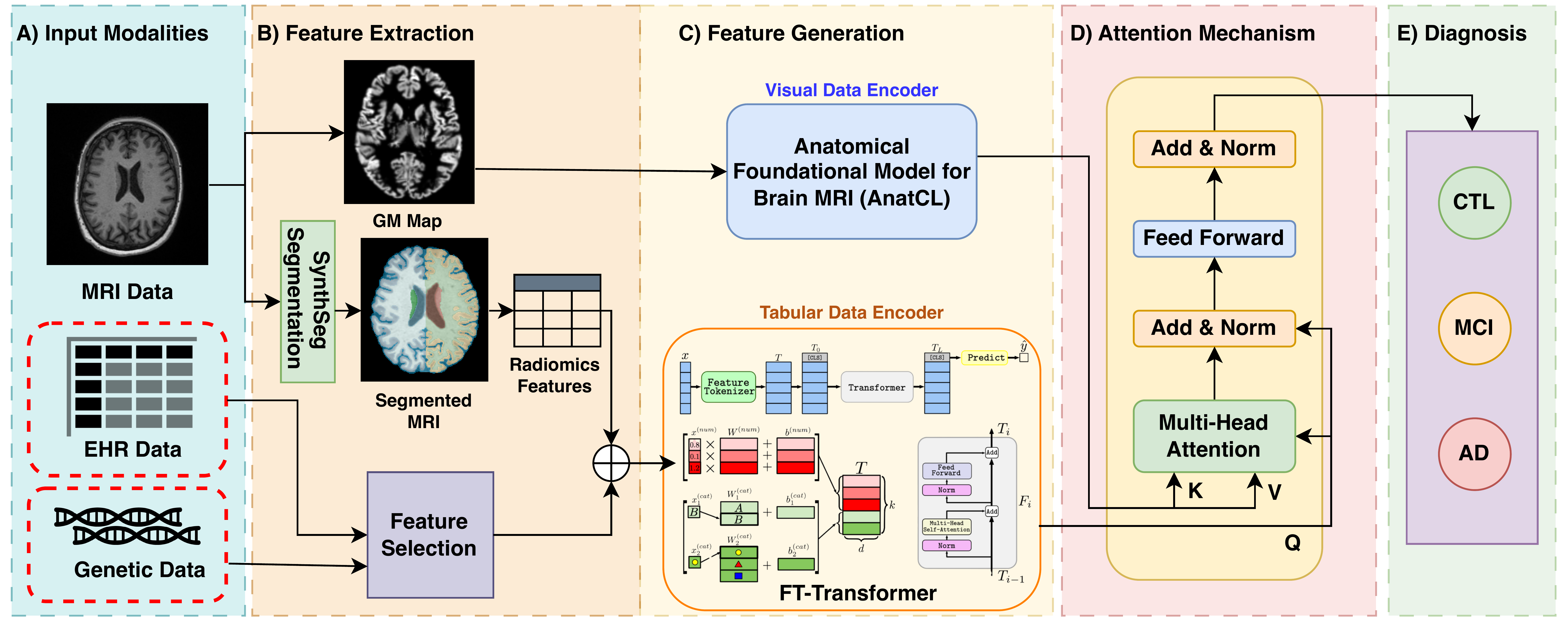}
    \caption{Multimodal diagnostic pipeline for dementia classification. MRI scans are processed to extract grey matter maps and radiomics features. Genetic and EHR data undergo feature selection. Grey matter is passed through AnatCL, while selected tabular features flow through the FT-Transformer. A cross-attention module integrates both representations before predicting dementia class (CTL, MCI, AD). Missing modalities such as EHR and genetic data are outlined in red to indicate their potential absence.}
    \label{fig:methodology}
\end{figure*}

The proposed framework processes MRI data by integrating genetic and clinical metadata to enhance accuracy, without requiring additional resources such as costly gene sequencing data at inference time. By effectively addressing the challenges of generalizability and interpretability, the proposed system demonstrates better real-world feasibility and outperforms existing methods. The key contributions of our work are as follows:

\begin{enumerate}
    \item \textbf{Comprehensive Multimodal Learning:} OmniBrain, an elegant multimodal model that fuses structural MRI, radiomics, cognitive scores, and gene expression data into a single framework, providing the most holistic modeling view of AD pathology.
    \item \textbf{Robust Missing-Modality Strategy:} Although training is performed on multimodal data, we employ a practical method combining \textit{Modality-Aware Attention Masking}, enabling the model to effectively handle missing modalities during inference and enhancing its robustness for real-world applications.
    \item \textbf{State-of-the-Art Performance and Transferability:} We achieve new SOTA results on ANMerge and demonstrate effective generalization to ADNI, even with incomplete inputs, reinforcing the framework’s real-world applicability.
\end{enumerate}

\section{Related Work}
\label{sec:related_work}

\textbf{Benchmarking AD Classification Results.}
The Alzheimer’s Disease Neuroimaging Initiative (ADNI) is widely used as a benchmark for Alzheimer’s classification. \citet{Zarei2024} applied 3D-CNN models combined with hippocampal and amygdala radiomics, reporting an accuracy of $84.4\%$ for AD vs.\ CTL, but only $\sim57\%$ for AD vs.\ MCI. \citet{Maddalena2022} used the ANMerge dataset, which includes MRI, gene expression, and clinical measures for over 1,700 participants. They fused MRI and gene expression data, achieving $94.6\%$ accuracy for AD vs.\ CTL, but performance dropped to $56\%$ for MCI vs.\ CTL. \citet{huang2025predicting} utilized graph optimal transport to explore molecular subtypes in ANMerge, though diagnostic classification was not the primary focus.

\textbf{Multimodal Approaches for Alzheimer’s Classification.}
Several works have explored combining modalities to better model AD pathology. The 3MT cross-attention transformer with modality dropout achieved $93.0\%$ accuracy (AUC 0.970) on ADNI/AIBL datasets \citep{Liu2023}. \citet{Henriquez2024} introduced a deep ensemble method that scored $98.8\%$ for CTL vs.\ MCI on a small cohort. MINDSETS integrated longitudinal MRI radiomics, transcriptomics, and cognitive scores to achieve $89.25\%$ for AD vs.\ VaD \citep{hassan2024mindsets}, though multiclass performance and reproducibility were limited. MADDi employed cross-modal attention to merge MRI, genetic, and clinical data and reached $96.9\%$ accuracy on ADNI without offering interpretability tools \citep{Golovanevsky2022}. \citet{Castellano2024} combined MRI and amyloid PET scans to enhance classification on OASIS-3, but the dependence on dual imaging restricted applicability.

\textbf{Handling Missing Modalities in Alzheimer’s Disease Models.}
Prior studies implemented generative imputation or modality-agnostic inference. 3MT used modality-dropout with cross-attention to manage incomplete inputs \citep{liu2022cascaded}. ITCFN co-attended MRI and clinical data to hallucinate PET imaging for MCI-conversion prediction \citep{hu2025itcfn}. Flex-MoE enabled dynamic routing through expert networks tailored to different modality combinations \citep{yun2024flexmoe}.

\textbf{Explainability in Alzheimer’s Models.}
Most models apply saliency-based techniques for interpretability. Grad-CAM and 3D Grad-CAM have been used to visualize spatial attention in CNNs \citep{Song2024}, while LIME has been applied for local surrogate explanations \citep{Adarsh2024}. \citet{FrontiersXAI2023} used dictionary learning and CNN ensembles to provide visual explanations. \citet{Alatrany2024} combined rule-based SVMs with SHAP values for tabular interpretability.

\noindent Existing approaches are constrained by one or more of : (i) limited modality coverage, (ii) weak cross-dataset generalization, (iii) fragile handling of missing modalities at inference, and (iv) scarce interpretability. Even methods explicitly designed for missing-modality settings are almost always trained and evaluated on the same cohort, leaving out-of-distribution robustness untested. \textit{OmniBrain} addresses these gaps in a single deployment-ready framework that raises accuracy, sustains performance on unseen datasets, and offers explainability, yielding a practical solution.

\begin{table}[ht]
\centering
\caption{Summary of key aspects covered in Alzheimer’s disease diagnosis papers. \checkmark = present, \xmark = not present.}
\label{tab:methodology_comparison}
\renewcommand{\arraystretch}{1.2}
\setlength{\tabcolsep}{3pt}
\resizebox{0.48\textwidth}{!}{%
\begin{tabular}{@{}lcccc@{}}
\toprule
\textbf{Paper} & \textbf{Multi-modal} & \textbf{Handling Missing Modality} & \textbf{Cross Dataset Validation} & \textbf{Explainability} \\
\midrule
Aderghal et al.\ (2018)~\cite{aderghal2018classification}       & \xmark & \xmark & \xmark & \xmark \\
Li et al.\ (2018)~\cite{li2018classification}                   & \xmark & \xmark & \xmark & \xmark \\
Pan et al.\ (2018)~\cite{pan2018multi}                          & \checkmark & \xmark & \xmark & \xmark \\
Shi et al.\ (2018)~\cite{shi2018multimodal}                     & \checkmark & \xmark & \xmark & \xmark \\
Zheng et al.\ (2018)~\cite{zheng2018identification}             & \xmark & \xmark & \xmark & \xmark \\
Backström et al.\ (2018)~\cite{backstrom2018efficient}          & \xmark & \xmark & \xmark & \xmark \\
Maddalena et al.\ (2022)~\cite{maddalena2022integrating}        & \checkmark & \xmark & \xmark & \xmark \\
Maddalena et al.\ (2023)~\cite{Maddalena2023}                   & \checkmark & \xmark & \xmark & \checkmark \\
Hassan et al.\ (2024)~\cite{hassan2024mindsets}                 & \checkmark & \xmark & \xmark & \checkmark \\
\rowcolor{lightblue} OmniBrain (Ours)                           & \checkmark & \checkmark & \checkmark & \checkmark \\
\bottomrule
\end{tabular}%
}
\end{table}

\section{Dataset}
\label{sec:Dataset}

Alzheimer’s disease (AD) typically progresses through three clinical stages: CTL $\rightarrow$ MCI $\rightarrow$ AD. CTL individuals exhibit no measurable cognitive decline, whereas MCI marks a transitional stage of mild yet noticeable impairment, which is particularly challenging to classify reliably. AD involves significant memory deficits and functional decline. Given the subtle differences between these groups, especially around MCI onset, accurate diagnosis remains clinically challenging.

\textbf{ANMerge Dataset.}
Our primary dataset, ANMerge, is a multimodal resource comprising 1,702 participants from European cohorts \cite{ANMerge}. It includes five key data types: clinical assessments (all participants), MRI scans (453), proteomics (680), gene expression (709), and genotyping (1,014). Notably, 239 individuals have complete data across all five modalities. The MRI subset consists of 453 participants with scans collected at up to three time points (0, 3, and 12 months), resulting in 1,067 MRI samples. For this study, focusing on MRI, genes, and clinical data, we identified 319 unique patients, totaling 919 samples. The \textit{Mini-Mental State Examination} (MMSE) assesses a global cognitive function, with lower scores indicating more severe impairment. The Apolipoprotein E (\textit{APOE}) gene, especially the E4 allele, significantly increases AD risk and accelerates disease onset, making APOE status an essential biomarker. The gene expression data covers over 16,000 post-processed genes (including whole-body genes).


\textbf{ADNI Dataset.}
As an external test set to measure our generalization to other datasets, we used the Alzheimer’s Disease Neuroimaging Initiative (ADNI), a large-scale, multi-phase study focused on identifying AD biomarkers \cite{ADNI}. We sampled 300 different MRI scans of patients (100 per class). ADNI provides a standardized MRI dataset, including 3D T1, FLAIR, diffusion, and functional MRI—from over 800 individuals (CTL, MCI, AD) in ADNI1. These openly available scans are widely applied in neuro-imaging research to explore disease trajectories and validate classification models.

\subsection{Data Pre-processing}

\begin{figure}[t!]
    \includegraphics[width=0.95\linewidth]{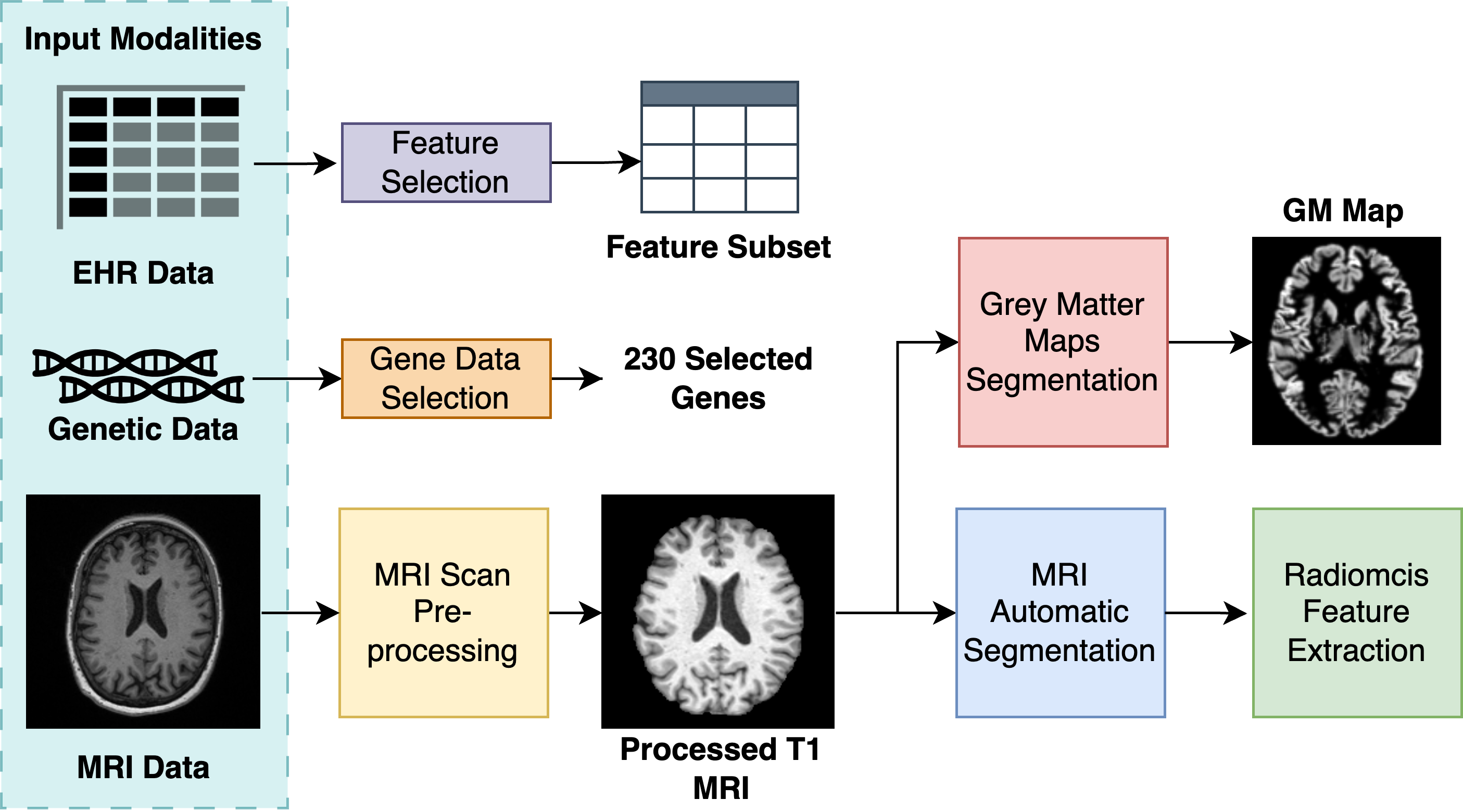}
    \caption{Overview of the multimodal pre-processing pipeline for MRI, genetic, and EHR data, including segmentation, feature selection.}
    \label{fig:full_dataset_processing}
\end{figure}

Figure \ref{fig:full_dataset_processing} outlines the multimodal dataset's pre-processing, feature extraction, and selection flow. Each data modality undergoes a dedicated processing stream. EHR data is filtered through feature selection to retain relevant clinical variables. Genetic data is narrowed down to a curated set of 230 genes using domain-guided selection. MRI data is first preprocessed, following two parallel paths: one for generating grey matter probability maps through tissue segmentation, and another for performing anatomical segmentation and extracting radiomics features. The next sections describe the details of each component.

\begin{figure}
    \centering
    \includegraphics[width=0.95\linewidth]{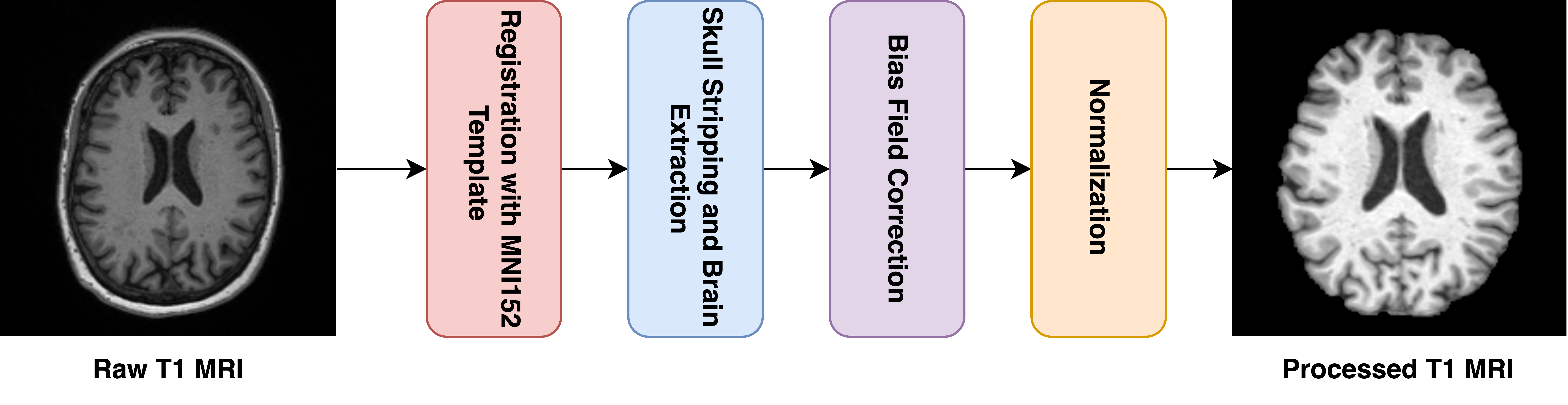}
    \caption{Pre-processing pipeline for MRI data for standardization, removal of artifacts, and normalization.}
    \label{fig:mri_preprocessing}
\end{figure}

\noindent
Working with neuroimaging data typically requires standardized pre-processing for consistency in downstream analyses. For ANMerge, the pipeline comprised four key stages as shown in Figure~\ref{fig:mri_preprocessing}): 
\begin{enumerate}
    \item Spatial registration with the MNI152 template (conventional template in brain registration) to ensure anatomical correspondence.
    \item Skull stripping and brain extraction to remove non-brain structures.
    \item Bias field correction using the N4 algorithm to mitigate intensity variations.
    \item Normalization signal intensities across scans.
\end{enumerate}
No preprocessing steps were required for the ADNI dataset, since it already includes comparable corrections.

Neither ANMerge nor ADNI provides ground truth segmentation masks for isolating brain structures. To address this, we employed \textit{SynthSeg} \cite{synthseg}, a pre-trained 3D U-Net capable of segmenting 32 brain regions, including the cortex, hippocampus, cerebellum, and brainstem—in both hemispheres. We selected this deep learning approach over traditional methods like FreeSurfer and FSL, as these may underestimate volumes in atrophied regions, potentially affecting analyses related to neurodegeneration \cite{sadil2024comparing}.

To validate SynthSeg’s outputs, we adopted three quality-control measures. First, a neuroradiologist reviewed a random subset of segmented scans balanced by class label (control, MCI, AD) to ensure anatomical plausibility of each region. Second, we systematically monitored SynthSeg’s built-in quality control (QC) metric to detect potential segmentation anomalies across all images. Third, despite the lack of pixel-wise ground truth masks, ANMerge provides volumetric reference measurements for each brain region, which we compared with SynthSeg-derived volumes. Strong concordance between these measurements further supported the accuracy of SynthSeg’s segmentations.

\subsection{Feature Extraction}

\textbf{Radiomics Features.} Radiomic features were extracted from 32 brain regions segmented by SynthSeg using PyRadiomics \cite{pyradiomics}. For each region, we computed 137 features encompassing first-order intensity statistics, shape descriptors, and texture metrics. In addition, 37 matrix-based texture features GLCM (gray-level co-occurrence), GLRLM (run-length), GLSZM (size zone), GLDM (dependence), and NGTDM (neighborhood difference) were derived to quantify intensity distributions, geometric properties, and heterogeneity. PyRadiomics applied standardized preprocessing—resampling to a common voxel size, intensity normalization, and gray-level discretization—to ensure consistent feature values across varying resolutions. These quantitative measures enable detection of subtle structural alterations in Alzheimer’s disease, notably in the hippocampus, cortex, and thalamus, often observable years before clinical onset.  

\textbf{Grey Matter (GM) Maps:}
To avoid relying solely on tabular representations and to harness deep learning’s capacity for spatial pattern recognition, we also extracted normalized GM maps with CAT12 where gray matter atrophy correlates strongly with disease progression \cite{cat}.  These voxel-level GM maps retain high-resolution anatomical detail, capturing localized atrophy patterns that might be overlooked in coarser representations. These subtle morphological changes are critical for distinguishing control, MCI, and AD groups by focusing on GM.


\subsection{Genetic Data}

ANMerge provides a rich repository of genetic data, including gene expressions and probes. Probes enable the detection and quantification of gene expression levels, formatted in a table with median-aggregated values from multiple probes targeting the same gene. This highlights patterns of gene expression that vary across samples, indirectly reflecting underlying genetic variations. In the context of AD, certain genetic variants are linked to disease risk and progression, making these data valuable for understanding the disease \cite{lunnon2013blood, oh2022alzheimer, adsp_gvc_2023}.

To use this data, we reviewed previous studies to identify expressions potentially relevant to AD. We began with a panel of Illumina probes from \cite{lunnon2013blood}, selecting genes showing strong diagnostic potential, along with top-ranked genes based on statistical significance. Recent studies \cite{oh2022alzheimer, adsp_gvc_2023} also identified biomarkers involved in AD pathogenesis. From these, we chose genes overlapping with ANMerge that are relevant for diagnosis. Altogether, these complementary sources yielded $230$ AD-related genes for our multimodality experiments.

\textbf{Statistical Feature Selection with F-score and p-value.} Although these genes have been linked to AD, not all showed strong associations in our dataset. To identify the most informative features from this high-dimensional set, we applied an ANOVA F-test \cite{fisher1925statistical} using \texttt{SelectKBest}. The F-score reflects how much a feature’s values differ between classes relative to within-class variation. Each score has a p-value indicating the likelihood that the difference occurred by chance. Features with p-values greater than 0.01 were discarded. We then computed a combined score, weighting normalized F-scores (70\%) and negative log-transformed p-values (30\%), to balance discriminative power and significance. The top-ranked \textit{139 genes} were selected for analysis. This has been done in folds to avoid label leakage between training and testing sets.

\section{Methodology}
\label{sec:method}
Our methodology constructs a multimodal system leveraging pre-trained, expert-level backbones for each modality. T1-weighted MRI scans are processed into GM maps to extract radiomics features. Concurrently, genetic and EHR data undergo feature selection to emphasize clinically significant markers. An anatomical foundation model (AnatCL or y-aware) encodes GM representations, while selected tabular features (radiomics, genetics, demographics) are embedded using the FT-Transformer. A cross-attention module fuses MRI and tabular embeddings into a unified latent space, enabling a classifier to predict dementia status (CTL, MCI, or AD). This design (see Figure~\ref{fig:methodology}) robustly accommodates missing data by treating absent modalities as nonexistent, ensuring consistent performance despite incomplete inputs.

\textbf{AnatCL} is an anatomical foundation model for brain MRI analysis, incorporating structural and demographic information via a weakly supervised contrastive framework~\cite{anatcl}. Trained on roughly 5,000 healthy T1-weighted MRIs (OpenBHB dataset), it has been evaluated across multiple clinical datasets, achieving SOTA performance on 12 neuroimaging tasks. By capturing subtle anatomical variations, AnatCL excels in early Alzheimer’s detection and serves as a robust backbone for classification models, offering generalization across diverse populations and MRI protocols.

\textbf{y-Aware InfoNCE} is a contrastive learning approach that integrates continuous metadata (e.g., age) into self-supervised learning for 3D brain MRIs~\cite{y-aware}. Negative samples are weighted by a Gaussian kernel, focusing the model on biologically relevant, metadata-informed differences. Pretrained on over 10,000 healthy T1 MRIs from multiple sources, it generalizes effectively across varied anatomical structures. Its age-sensitive features are especially pertinent for Alzheimer’s, where atrophy correlates with aging, making it a strong backbone for clinical tasks with limited data.

The \textbf{FT-Transformer (Feature Tokenizer + Transformer)} targets tabular data; it unifies categorical and numerical features into an embedded sequence prepended by a [CLS] token, subsequently processed via transformer layers with multi-head self-attention. Numerical features \( x_j^{\text{num}} \) are embedded as \( T_j^{\text{num}} = b_j + x_j \cdot W_j \), while categorical features \( x_j^{\text{cat}} \) use embedding lookups \( T_j^{\text{cat}} = b_j + W_j[x_j^{\text{cat}}] \), with learnable parameters \( W_j, b_j \). The transformer applies stacked layers and generates predictions from the final [CLS] token representation.
\begin{equation}
\label{Eq.ft1}
T_l = \text{FFN}\bigl(\text{LayerNorm}(T_{l-1} + \text{MHSA}(\text{LayerNorm}(T_{l-1})))\bigr),
\end{equation}

This architecture captures complex inter-feature interactions, essential for Alzheimer’s studies involving imaging metrics (e.g., hippocampal volume) and genetic markers (e.g., APOE variants). Its permutation invariance and robustness to missing data make FT-Transformer particularly effective for predicting Alzheimer’s progression from multimodal tabular inputs.

\textbf{Missing Modality.}
In real‑world clinical workflows, multimodal records are frequently incomplete, demanding models that remain reliable when one or more inputs are absent. We therefore adopt a \emph{Modality-Aware Attention Masking } scheme that handles missing data \cite{wu2024deepmultimodallearningmissing}. Each source—structural MRI, radiomics, or tabular signals such as gene expression and cognitive scores—is first encoded into its own latent representation. A cross‑modal attention layer then weighs the available representations according to their contextual importance, while masks missing modalities, integrating only the information that is actually present at inference time. By design, the network neither reconstructs nor imputes unseen modalities; instead, it emphasises the evidence at hand and suppresses empty channels, preserving interpretability while avoiding error propagation.

To ensure this behaviour generalises beyond curated datasets, we randomly drop specific modalities by masking them during training, forcing the attention mechanism to learn adaptive fusion weights rather than depend on any single stream. This strategy delivers stable performance across diverse input configurations, effectively narrowing the gap between controlled research settings and practical deployment environments where missing data is commonplace.

\textbf{Explainability.} To interpret our model, we applied Gradient‐weighted Class Activation Mapping (Grad‐CAM) for visual data (Grey Matter) and SHAP (SHapley Additive exPlanations) for tabular data (radiomics, gene‑expression, and clinical metadata).  \textit{Grad‐CAM} ~\cite{selvaraju2017grad} uses gradients of the target class score  with respect to the final convolutional feature maps  to produce a coarse localization map. \textit{SHAP} is a model‑agnostic additive feature‑attribution framework that assigns each feature \(i\) a Shapley value \(\phi_i\) by averaging its marginal contribution across all coalitions \(S\subseteq F\setminus\{i\}\) \cite{lundberg2017unified}. SHAP excels on tabular data by handling heterogeneous features, capturing interactions via coalition values, and leveraging efficient algorithms (e.g., TreeSHAP) that scale polynomially with the number of features and samples.

\section{Experimental Setup}
\label{sec:setup}

Our goal is to develop and validate a framework that integrates MRI, genetic, and clinical metadata for AD classification, achieving high accuracy and strong generalization without requiring the presence of all modalities in use during inference. Figure~\ref{fig:methodology} summarizes the overall design: (1) grey-matter images are encoded via pre-trained backbone models (AnatCL or y-Aware InfoNCE), (2) radiomics, gene expression, and clinical metadata are fed into the FT-Transformer, and (3) a cross-attention module fuses these representations to classify individuals into AD, MCI, or CTL. We fine-tune the pre-trained components using four cross-attention heads to limit overfitting.

\paragraph{ANMerge Training Details.}
We focus primarily on the ANMerge dataset, a 5-fold group-based cross-validation that splits data at the patient level, reserving 20\% of patients as a test subset in each fold to avoid data leakage. We train on a single NVIDIA A6000 GPU (48 GB VRAM). For tabular inputs, categorical variables are label-encoded, continuous variables are scaled via a standard scaler, and missing entries are imputed with mean or mode values based on each class. We train for up to 50 epochs, applying early stopping (patience of 5) to prevent overfitting. The Adam optimizer uses a learning rate of 0.00001, a weight decay of \(5 \times 10^{-4}\), and a batch size of 64. We employ a weighted focal loss to address class imbalance, where each class weight is inversely proportional to its frequency.

\paragraph{ADNI Testing and Generalization.}
To evaluate real-world applicability, we also test on ADNI. ADNI lacks gene and extended metadata, so we employ the same cross-attention architecture but apply the selective attention fusion missing-modality method at inference. Specifically, we train on all ANMerge samples (totaling 919) and then test on ADNI (300 scans), making ADNI roughly 25\% of the combined dataset. This procedure demonstrates the framework’s robustness to incomplete inputs and different data distributions.

\paragraph{Comparative Modalities and Methods.}
We evaluate each modality individually—MRI-only or tabular-only—and then explore their combinations. We compare two pre-trained backbones for MRI (AnatCL and y-Aware). For tabular data (radiomics, genes, metadata), we contrast the FT-Transformer approach with XGBoost, hyperparameter-tuned via Optuna. These comparisons provide a thorough assessment of how various data modalities and modeling choices contribute to overall classification performance.

\section{Results \& Discussion}
\label{sec:results}


\newcommand{\filledcirc}{\ding{108}}
\newcommand{\opencirc}  {\ding{109}}

\sisetup{detect-mode,detect-family,table-format=1}
\begin{table}[ht]
\centering
\caption{Mean ± SD performance (\%) for ANMerge Dataset.
{\color{mildgreen}Light-green} rows = \emph{missing-modality} runs.
\textbf{\underline{Bold underlined}} cells mark the best results.}
\label{tab:anmerge_performance}

\resizebox{0.48\textwidth}{!}{%
\begin{tabular}{cccc  ccc  ccc}
\toprule
\rowcolor{lyheader}\multicolumn{4}{c}{\textbf{Data}} &
\multicolumn{3}{c}{\textbf{Models}} & \multicolumn{3}{c}{\textbf{Metrics}}\\
\cmidrule(lr){1-4}\cmidrule(lr){5-7}\cmidrule(l){8-10}
\textbf{Rad} & \textbf{GM} & \textbf{Genes} & \textbf{Meta} &
\textbf{AnatCL} & \textbf{Y-aware} & \textbf{FT-Tr} &
\textbf{Accuracy} & \textbf{Recall} & \textbf{F1-Score}\\
\midrule
\rowcolor{lrline}\multicolumn{10}{c}{Single-modality (1 signal)}\\
\midrule
\filledcirc & \opencirc & \opencirc & \opencirc & \opencirc & \opencirc & \filledcirc & 73.50$\pm$1.51 & 74.22$\pm$2.53 & 74.50$\pm$1.97\\
\opencirc & \filledcirc & \opencirc & \opencirc & \opencirc & \filledcirc & \opencirc & 69.53$\pm$2.55 & 68.68$\pm$2.11 & 69.52$\pm$1.28\\
\opencirc & \filledcirc & \opencirc & \opencirc & \filledcirc & \opencirc & \opencirc & 73.18$\pm$2.46 & 74.04$\pm$1.79 & 73.17$\pm$2.28\\
\midrule
\rowcolor{lrline}\multicolumn{10}{c}{Dual-modality (2 signals)}\\
\midrule
\filledcirc & \opencirc & \filledcirc & \opencirc & \opencirc & \opencirc & \filledcirc & 84.50$\pm$1.37 & 84.04$\pm$2.56 & 84.50$\pm$2.78\\
\rowcolor{lightgreen}\filledcirc & \opencirc & \filledcirc & \opencirc & \opencirc & \opencirc & \filledcirc & 81.72$\pm$2.52 & 81.52$\pm$2.23 & 81.72$\pm$1.30\\
\filledcirc & \filledcirc & \opencirc & \opencirc & \opencirc & \filledcirc & \filledcirc & 82.04$\pm$2.18 & 81.92$\pm$2.52 & 82.04$\pm$2.04\\
\filledcirc & \filledcirc & \opencirc & \opencirc & \filledcirc & \opencirc & \filledcirc & 79.87$\pm$1.64 & 80.29$\pm$2.39 & 79.87$\pm$1.04\\
\midrule
\rowcolor{lrline}\multicolumn{10}{c}{Triple-modality (3 signals)}\\
\midrule
\filledcirc & \opencirc & \filledcirc & \filledcirc & \filledcirc & \opencirc & \filledcirc & 88.54$\pm$1.08 & 88.46$\pm$2.74 & 88.54$\pm$2.70\\
\rowcolor{lightgreen}\filledcirc & \opencirc & \filledcirc & \filledcirc & \filledcirc & \opencirc & \filledcirc & 85.48$\pm$1.54 & 84.60$\pm$2.37 & 85.47$\pm$1.04\\
\filledcirc & \filledcirc & \filledcirc & \opencirc & \opencirc & \filledcirc & \filledcirc & 85.12$\pm$1.25 & 84.46$\pm$2.15 & 85.12$\pm$1.56\\
\rowcolor{lightgreen}\filledcirc & \filledcirc & \filledcirc & \opencirc & \opencirc & \filledcirc & \filledcirc & 82.51$\pm$1.88 & 81.82$\pm$1.61 & 82.51$\pm$1.13\\
\midrule
\rowcolor{lrline}\multicolumn{10}{c}{Quadruple-modality (all 4 signals)}\\
\midrule
\filledcirc & \filledcirc & \filledcirc & \filledcirc & \opencirc & \filledcirc & \filledcirc & 90.87$\pm$2.39 & 90.53$\pm$2.52 & 90.87$\pm$1.12\\
\rowcolor{lightgreen}\filledcirc & \filledcirc & \filledcirc & \filledcirc & \opencirc & \filledcirc & \filledcirc & 87.14$\pm$1.68 & 87.65$\pm$1.77 & 87.14$\pm$2.40\\
\textbf{\filledcirc} & \textbf{\filledcirc} & \textbf{\filledcirc} & \textbf{\filledcirc} &
\filledcirc & \opencirc & \filledcirc &
\textbf{\underline{92.15$\pm$2.37}} & \textbf{\underline{91.12$\pm$2.09}} & \textbf{\underline{92.15$\pm$1.70}}\\
\rowcolor{lightgreen}\filledcirc & \filledcirc & \filledcirc & \filledcirc & \filledcirc & \opencirc & \filledcirc & 89.79$\pm$2.77 & 90.95$\pm$2.22 & 89.55$\pm$1.57\\
\bottomrule
\end{tabular}}
\end{table}


\begin{table}[ht]
\centering
\caption{Mean ± SD performance (\%) for ADNI Dataset.
{\color{mildgreen}Light-green} rows = \emph{missing-modality} runs.
\textbf{\underline{Bold underlined}} cells mark the best results.}
\label{tab:adni_performance}

\resizebox{0.48\textwidth}{!}{%
\begin{tabular}{cccc  ccc  ccc}
\toprule
\rowcolor{lyheader}\multicolumn{4}{c}{\textbf{Data}} &
\multicolumn{3}{c}{\textbf{Models}} & \multicolumn{3}{c}{\textbf{Metrics}}\\
\cmidrule(lr){1-4}\cmidrule(lr){5-7}\cmidrule(l){8-10}
\textbf{Rad} & \textbf{GM} & \textbf{Genes} & \textbf{Meta} &
\textbf{AnatCL} & \textbf{Y-aware} & \textbf{FT-Tr} &
\textbf{Accuracy} & \textbf{Recall} & \textbf{F1-Score}\\
\midrule
\rowcolor{lrline}\multicolumn{10}{c}{Single-modality (1 signal)}\\
\midrule
\filledcirc & \opencirc & \opencirc & \opencirc & \opencirc & \opencirc & \filledcirc & 53.15$\pm$2.70 & 53.52$\pm$2.75 & 53.15$\pm$1.19\\
\opencirc & \filledcirc & \opencirc & \opencirc & \opencirc & \filledcirc & \opencirc & 49.14$\pm$2.88 & 48.41$\pm$2.74 & 49.13$\pm$2.36\\
\opencirc & \filledcirc & \opencirc & \opencirc & \filledcirc & \opencirc & \opencirc & 50.87$\pm$2.26 & 50.47$\pm$2.50 & 50.87$\pm$1.96\\
\midrule
\rowcolor{lrline}\multicolumn{10}{c}{Dual-modality (2 signals)}\\
\midrule
\rowcolor{lightgreen}\filledcirc & \opencirc & \filledcirc & \opencirc & \opencirc & \opencirc & \filledcirc & 58.93$\pm$2.63 & 59.23$\pm$1.99 & 58.93$\pm$2.85\\
\filledcirc & \filledcirc & \opencirc & \opencirc & \opencirc & \filledcirc & \filledcirc & 57.24$\pm$2.11 & 56.43$\pm$2.36 & 57.23$\pm$2.89\\
\filledcirc & \filledcirc & \opencirc & \opencirc & \filledcirc & \opencirc & \filledcirc & 55.19$\pm$2.71 & 54.42$\pm$1.52 & 55.18$\pm$2.83\\
\midrule
\rowcolor{lrline}\multicolumn{10}{c}{Triple-modality (3 signals)}\\
\midrule
\rowcolor{lightgreen}\filledcirc & \opencirc & \filledcirc & \filledcirc & \filledcirc & \opencirc & \filledcirc & 65.19$\pm$1.91 & 64.46$\pm$2.42 & 65.18$\pm$1.07\\
\rowcolor{lightgreen}\filledcirc & \filledcirc & \filledcirc & \opencirc & \opencirc & \filledcirc & \filledcirc & 62.07$\pm$2.02 & 61.36$\pm$2.19 & 62.07$\pm$1.28\\
\midrule
\rowcolor{lrline}\multicolumn{10}{c}{Quadruple-modality (all 4 signals)}\\
\midrule
\rowcolor{lightgreen}\filledcirc & \filledcirc & \filledcirc & \filledcirc & \opencirc & \filledcirc & \filledcirc & 68.78$\pm$1.18 & 69.98$\pm$2.07 & 68.74$\pm$2.40\\
\rowcolor{lightgreen}\filledcirc & \filledcirc & \filledcirc & \filledcirc & \filledcirc & \opencirc & \filledcirc & \textbf{\underline{70.37$\pm$2.68}} & \textbf{\underline{69.98$\pm$1.63}} & \textbf{\underline{70.37$\pm$1.24}}\\
\bottomrule
\end{tabular}}
\end{table}

Table~\ref{tab:anmerge_performance} reports the performance of all models—both single-modality and incremental multimodal fusion—evaluated in the in-domain scenario (trained and tested on ANMerge). Table~\ref{tab:adni_performance} summarizes results for the cross-domain scenario, where models trained on ANMerge are evaluated via zero-shot inference on the ADNI dataset. The \textit{missing-modality} strategy (highlighted in green) is applied to handle the absence of gene expression and/or clinical data in the ADNI experiments.

A radiomics-only baseline that uses FT-Transformer achieves $73.5\%$ accuracy on ANMerge but drops sharply to $53.1\%$ on ADNI, showing poor generalization. Introducing voxel-level GM representations (Y-aware or AnatCL) improves accuracy by $2$–$3\%$ in-domain and $\sim4\%$ cross-domain. AnatCL slightly outperforms Y-aware, indicating the added value of contrastive anatomical pretraining.

Incorporating gene expression or clinical metadata individually yields $4$–$8\%$ gains, and using both together further elevates performance to $88.5\%$ on ANMerge and $65.2\%$ on ADNI. This confirms that clinical signals offer essential, non-redundant information complementary to imaging. Models trained with modality dropout demonstrate resilience to missing inputs, where the accuracy only drops by 2-3\%  compared to the full multimodal input, which is around a 15\% gain in accuracy over training using MRI and radiomics inputs only. On ANMerge, in-domain accuracy degradation remains under $3.5\%$, while on ADNI, these models achieve $68.8\%$ accuracy, significantly surpassing the radiomics-only baseline by over $15\%$. This underscores the effectiveness of cross-modal attention in compensating for absent channels.

Our proposed framework, which integrates AnatCL for imaging and FT-Transformer for tabular data under a missing-modality strategy, achieves the highest accuracy: $92.15\%$ on ANMerge and $70.37\%$ on ADNI. This marks a substantial improvement of over $17.2\%$ compared to the best unimodal baseline, affirming our multimodal approach's superior results and generalization power over conventional MRI-centric methods.
\begin{figure*}[t]
  \centering
  \resizebox{0.8\textwidth}{!}{
    \begin{minipage}[b]{0.31\linewidth}\centering
      \includegraphics[width=\linewidth]{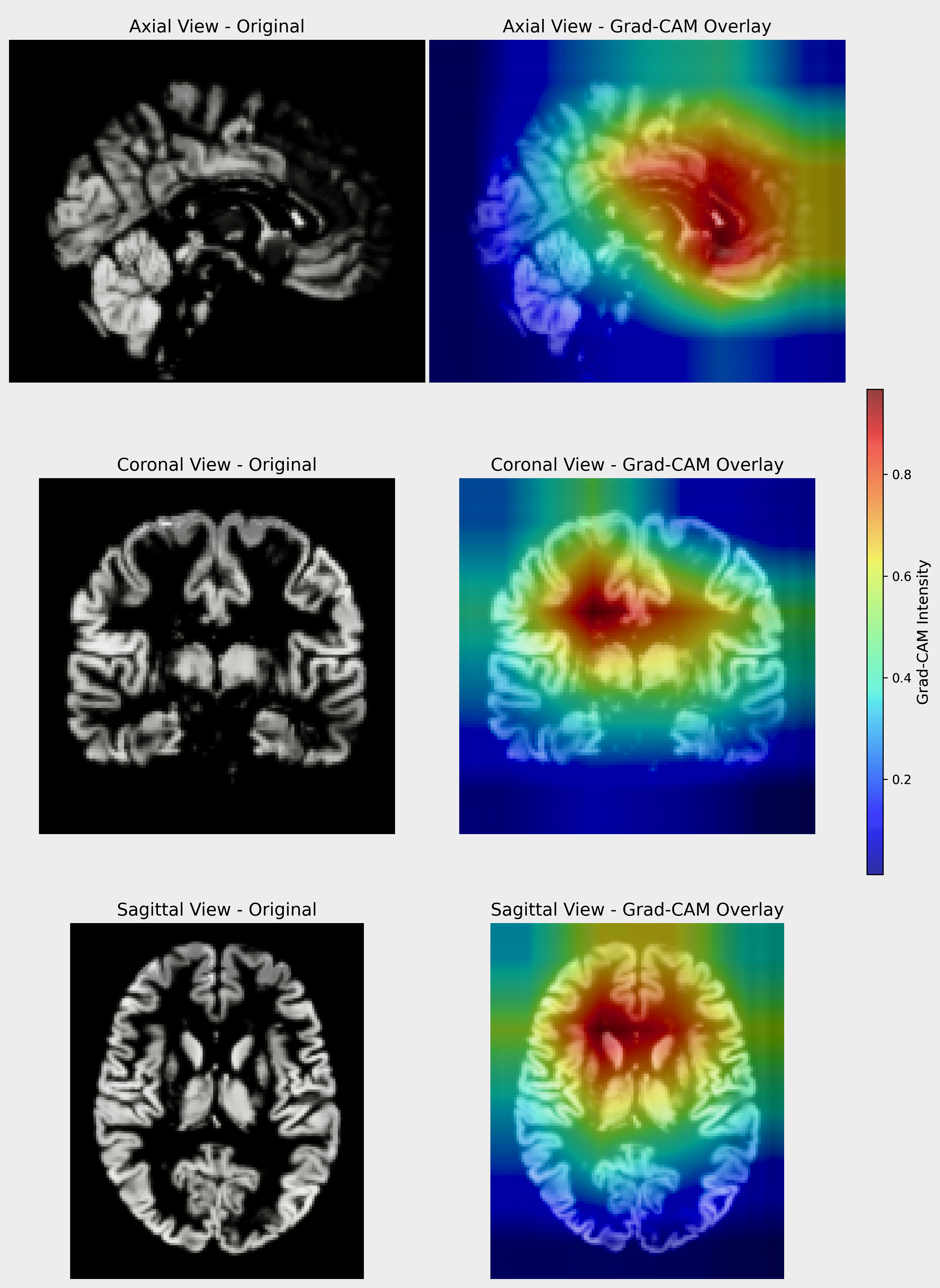}\par
      \vspace{1pt}\small (a) GradCam Visualization for (AD)
    \end{minipage}\hfill
    \begin{minipage}[b]{0.31\linewidth}\centering
      \includegraphics[width=\linewidth]{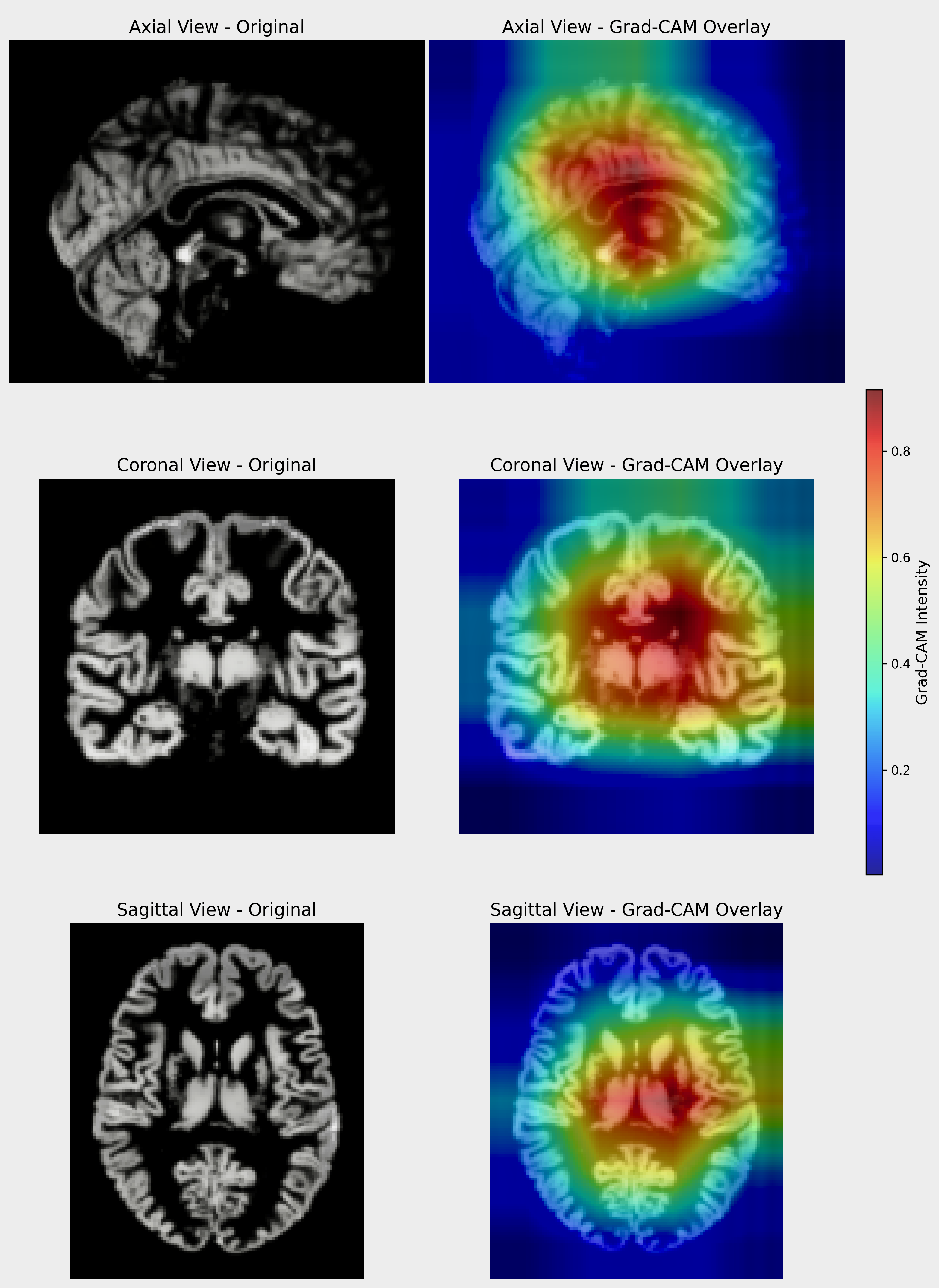}\par
      \vspace{1pt}\small (b) GradCam Visualization for (CTL)
    \end{minipage}\hfill
    \begin{minipage}[b]{0.31\linewidth}\centering
      \includegraphics[width=\linewidth]{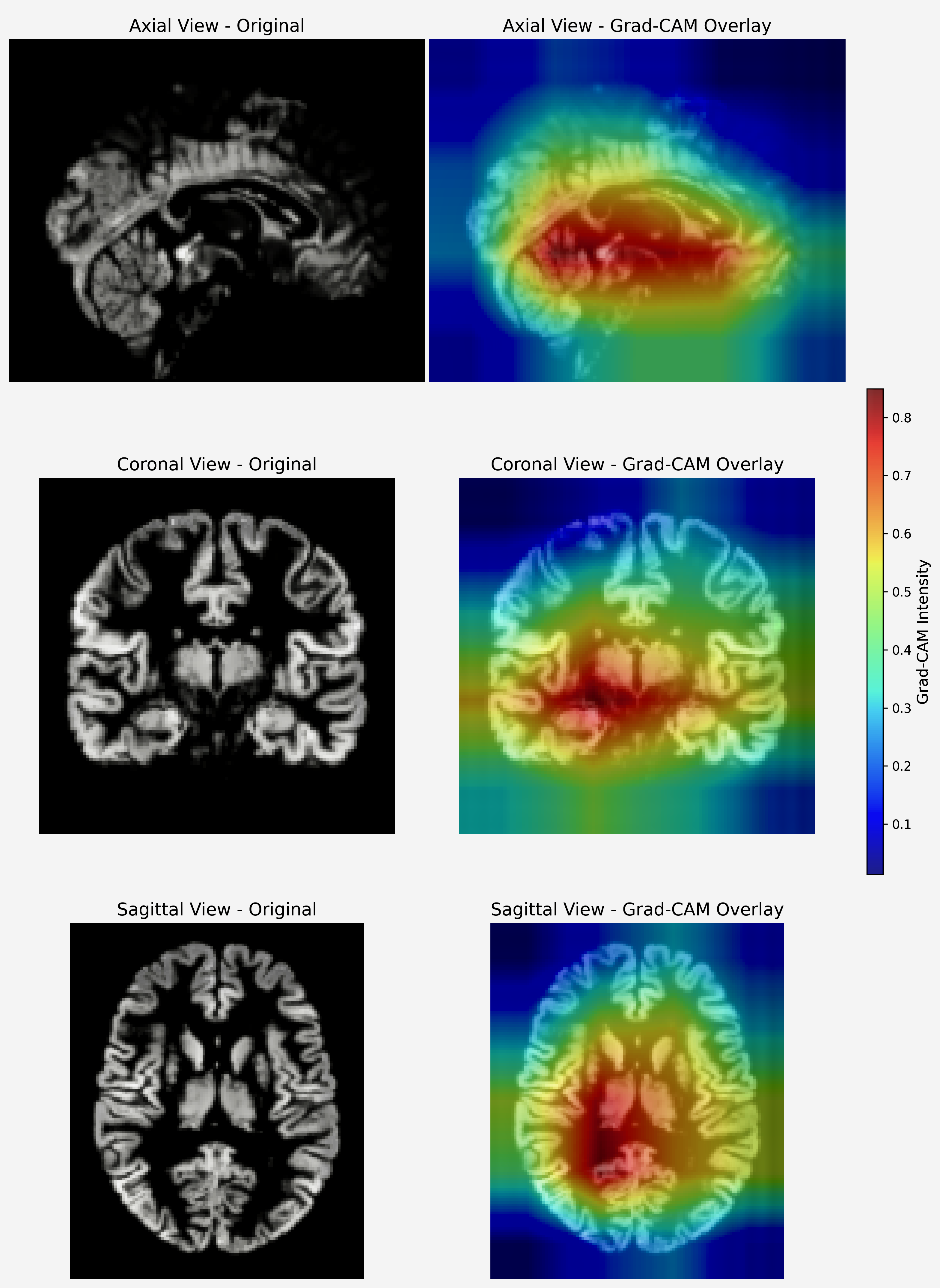}\par
      \vspace{1pt}\small (c) GradCam Visualization for (MCI)
    \end{minipage}%
  }
  \caption{Grad-CAM attention overlays in axial, coronal and sagittal views for
  (a) AD, (b) CTL and (c) MCI, showing the network’s consistent focus on
  medial-temporal and subcortical regions relevant to Alzheimer’s pathology.}
  \label{fig:gradcam_three}
\end{figure*}

\begin{table}[ht]
\centering
\caption{Performance (\%) of recent methods on various datasets (ANMerge \& ADNI) reporting Accuracy, Recall, and F1‐score.}
\label{tab:comparison_acc_recall_f1}
\setlength{\tabcolsep}{2pt}%
\resizebox{0.48\textwidth}{!}{%
\begin{tabular}{@{}lccccc@{}}   
\toprule
\textbf{Method} & \textbf{Dataset} & \textbf{Feats.} & \textbf{Acc} & \textbf{Recall} & \textbf{F1} \\
\midrule
\multicolumn{6}{c}{\textit{AD vs.\ CTL}} \\[1pt]
Aderghal et al.\ (2018)~\cite{aderghal2018classification} & ADNI    & MRI, DTI           & 92.5 & 94.7 & --   \\
Li et al.\ (2018)~\cite{li2018classification}             & ADNI    & MRI                & 89.5 & 87.9 & --   \\
Shi et al.\ (2018)~\cite{shi2018multimodal}               & ADNI    & MRI, PET           & 97.1 & 95.9 & --   \\
Maddalena et al.\ (2022)~\cite{maddalena2022integrating}  & ADNI    & MRI, GE            & 94.6 & 72.2 & 78.7 \\
Maddalena et al.\ (2023)~\cite{Maddalena2023}             & ANMerge & MRI, GE            & 87.4 & 87.2 & 87.2 \\
Hassan et al.\ (2024)~\cite{hassan2024mindsets}           & ANMerge & MRI, GE, Meta      & 99.35 & 99.91 & 99.31 \\
\rowcolor{lightblue} OmniBrain (Ours)                     & ANMerge & MRI, GE, Meta      & 99.17 & 98.48 & 99.29 \\
\midrule
\multicolumn{6}{c}{\textit{AD vs.\ MCI}} \\[1pt]
Aderghal et al.\ (2018)~\cite{aderghal2018classification} & ADNI    & MRI, DTI           & 85.0 & 93.7 & --   \\
Zheng et al.\ (2018)~\cite{zheng2018identification}       & ADNI    & MRI                & 73.8 & 64.1 & --   \\
Maddalena et al.\ (2022)~\cite{maddalena2022integrating}  & ADNI    & MRI, GE            & 91.5 & 39.4 & 44.8 \\
Maddalena et al.\ (2023)~\cite{Maddalena2023}             & ANMerge & MRI, GE            & 72.2 & 72.0 & 71.0 \\
Hassan et al.\ (2024)~\cite{hassan2024mindsets}           & ANMerge & MRI, GE, Meta      & 90.69 & 90.69 & 90.68 \\
\rowcolor{lightblue} OmniBrain (Ours)                     & ANMerge & MRI, GE, Meta      & 94.83 & 94.04 & 93.55 \\
\midrule
\multicolumn{6}{c}{\textit{MCI vs.\ CTL}} \\[1pt]
Aderghal et al.\ (2018)~\cite{aderghal2018classification} & ADNI    & MRI, DTI           & 80.0 & 92.8 & --   \\
Li et al.\ (2018)~\cite{li2018classification}             & ADNI    & MRI                & 73.8 & 86.6 & --   \\
Shi et al.\ (2018)~\cite{shi2018multimodal}               & ADNI    & MRI, PET           & 87.2 & 97.9 & --   \\
Maddalena et al.\ (2022)~\cite{maddalena2022integrating}  & ADNI    & MRI                & 63.6 & 73.2 & 71.7 \\
Maddalena et al.\ (2023)~\cite{Maddalena2023}             & ANMerge & MRI, GE            & 77.8 & 79.0 & 78.6 \\
Hassan et al.\ (2024)~\cite{hassan2024mindsets}           & ANMerge & MRI, GE, Meta      & 86.45 & 86.23 & 84.23 \\
\rowcolor{lightblue} OmniBrain (Ours)                     & ANMerge & MRI, GE, Meta      & 91.24 & 90.82 & 89.65 \\
\bottomrule
\end{tabular}%
} 
\end{table}

OmniBrain’s performance, highlighted in blue in Table~\ref{tab:comparison_acc_recall_f1}, underscores the strength of our multimodal fusion approach—combining MRI, gene expression, and lightweight metadata—to deliver state-of-the-art (SOTA) or near-SOTA results across all three binary Alzheimer’s classification tasks on the challenging ANMerge dataset. Specifically, it achieves 99.17\% accuracy and 99.29\% F1-score on the relatively straightforward AD vs.\ CTL task, closely aligning with the best existing benchmark that also utilizes the same modalities. More notably, OmniBrain exhibits substantial improvements on the more challenging AD vs.\ MCI and MCI vs.\ CTL tasks, attaining 94.83\% and 91.24\% accuracy, respectively—representing consistent gains of 4–6 percentage points over all prior models. Importantly, OmniBrain maintains a well-balanced precision–recall trade-off, with consistently high recall and F1-scores across tasks, suggesting that the model does not merely memorize late-stage disease cues but robustly learns the underlying trajectory of neurodegeneration. This highlights its potential not only for accurate classification but also for practical deployment in early diagnosis and longitudinal monitoring scenarios.

\begin{table}[ht]
\centering
\caption{Performance analysis of OmniBrain multimodal on ADNI Dataset in test mode only.}
\label{tab:omnibrain_dataset_gap}
\resizebox{0.48\textwidth}{!}{%
\begin{tabular}{@{}lccccc@{}}
\toprule
\textbf{Task} & \textbf{Accuracy\%} & \textbf{Recall \%} & \textbf{F1\%} & \textbf{$\Delta$Accuracy\%} & \textbf{$\Delta$F1\%} \\
\midrule
AD vs.\ CTL    & 86.29 & 84.69 & 85.39 & 13.88$\downarrow$ & 12.90 $\downarrow$ \\
\midrule
AD vs.\ MCI    & 82.55 & 80.87 & 81.45 & 13.28$\downarrow$ & 12.10 $\downarrow$ \\
\midrule
MCI vs.\ CTL   & 79.47 & 78.11 & 77.10 & 12.77$\downarrow$ & 11.55 $\downarrow$ \\
\bottomrule
\end{tabular}%
}
\end{table}


Table \ref{tab:omnibrain_dataset_gap} summarizes OmniBrain’s cross-dataset generalizability by testing a model trained only on ANMerge against the independent ADNI cohort. Performance remains high—85.29\% accuracy (85.39\% F1) for AD vs CTL, 81.55\% accuracy (80.45\% F1) for AD vs MCI, and 78.47\% accuracy (77.10\% F1) for MCI vs CTL—yet each task shows an absolute decrease ($\Delta$) of 11–13 percentage points relative to the corresponding ANMerge results. These drops are expected, given differences in participant demographics, imaging protocols, and feature distributions between the two datasets.

\paragraph{Model Explainability}


\paragraph{Visual Features via Grad-CAM:} We generated Grad‑CAM maps on sagittal, coronal, and axial slices and quantified each region’s attention intensity. Peak activations consistently align with the hippocampus, parahippocampal gyrus, and thalamus—the very loci neuroradiologists interrogate when grading medial‑temporal atrophy and subcortical integrity. Hippocampal saliency is maximal in AD, attenuates in MCI, and is lowest in CTL, while thalamic and parahippocampal signals remain strong yet exhibit subtle class‑specific shifts. This near‑perfect overlap between model attention and established pathological targets confirms the network’s diagnostic fidelity and enhances its interpretability for clinical deployment.

\begin{figure}[t!]
    \centering
    \includegraphics[width=0.9\linewidth]{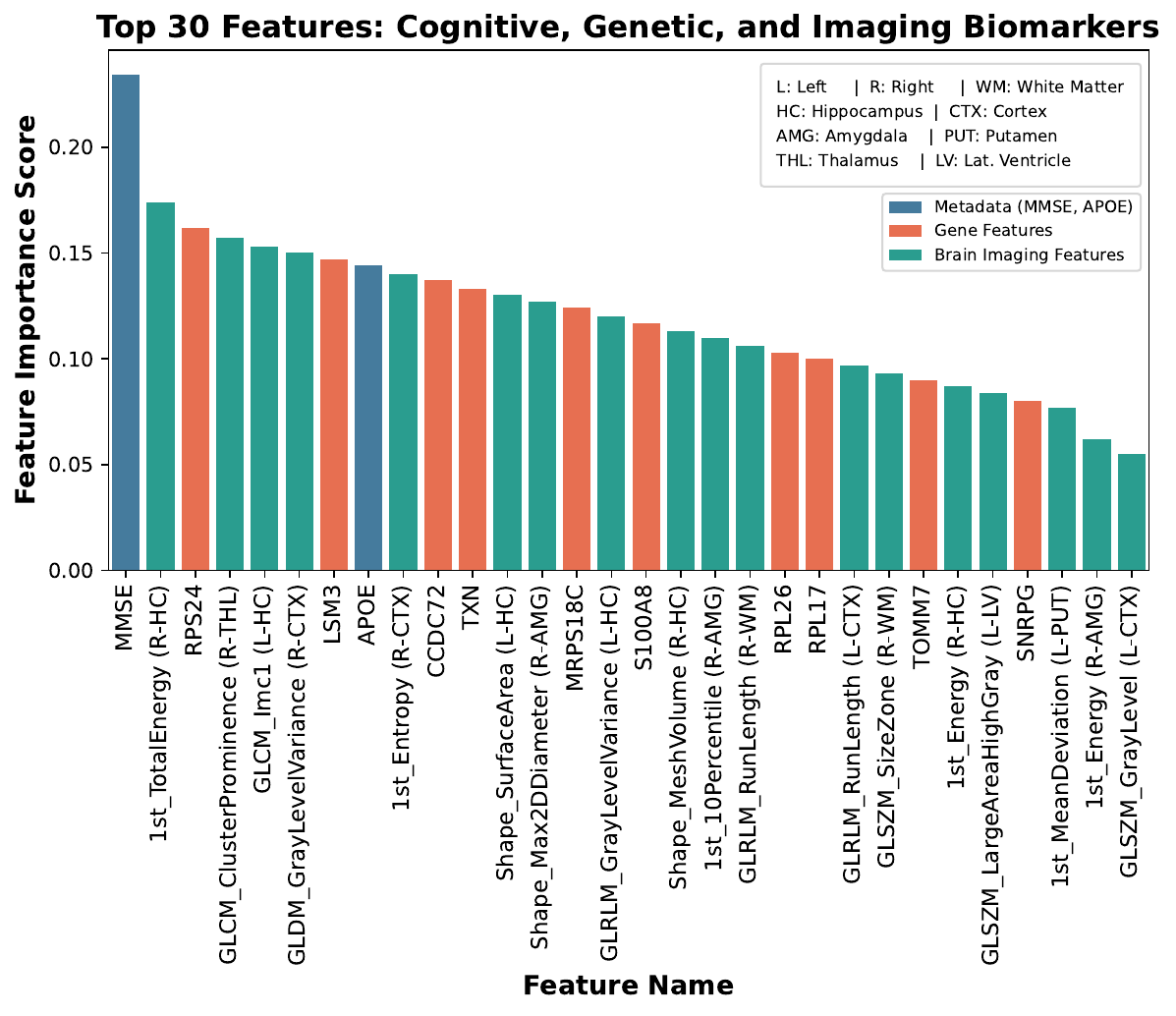}
    \caption{Axial slices of T1-weighted MRI showing processed images and segmented GM and WM for CTL, MCI, and AD groups.}
    \label{fig:feature_importance}
\end{figure} 

\paragraph{Tabular Features via SHAP.}  
Figure~\ref{fig:feature_importance} presents SHAP scores for the combined radiomics, gene‑expression, and clinical metadata features, establishing a biologically meaningful ranking of AD markers. \textbf{Clinical metadata}  \textit{MMSE} dominates the importance spectrum (SHAP\,$\approx$\,0.23), with \textit{APOE} also in the top decile—consistent with established cognitive and genetic risk factors in AD diagnosis and highlights their importance in clinical use for AD classification. For \textbf{Gene expression}, our feature importance analysis shows that eight of the top ten genes most strongly associated with Alzheimer’s disease prediction have previously been linked to oxidative stress, mitochondrial dysfunction, or ribosomal dysregulation in AD \cite{lunnon2013blood, oh2022alzheimer, adsp_gvc_2023}, , while the remaining two (\texttt{LSM3}, \texttt{CCDC72}) have not been previously linked to Alzheimer’s disease in the literature and need further investigation into their biological roles as potential novel biomarkers for AD. \textbf{Radiomics}  
Key imaging metrics (first‑order energy, entropy, mean deviation; GLCM Cluster Prominence; GLDM Gray Level Variance) derive predominantly from the hippocampus, amygdala, cortex, and thalamus, which represent the most important and relevant part of the brain for AD, which concords with our CNN’s Grad‑CAM saliency maps.
\section{Conclusion}
\label{sec:conclusion}

We introduced \textit{OmniBrain}, a powerful multimodal framework integrating voxel-level \emph{grey-matter maps}, radiomics from 32 SynthSeg-derived brain regions, blood-based \emph{gene-expression} profiles, and routine \emph{clinical metadata} via a cross-attention fusion block. Imaging cues are encoded by foundation models (AnatCL or \hbox{$y$-Aware InfoNCE}), while an FT-Transformer processes tabular features. \emph{Modality-Aware Attention Masking} during training dynamically teaches the model to re-weight evidence, enabling inference with any subset of inputs. OmniBrain achieves \textit{92.15\% $\pm$ 2.4} accuracy on the in-domain ANMerge cohort, setting new state-of-the-art benchmarks in the challenging AD vs. MCI and MCI vs. CTL tasks by up to \textit{+4.8\%}. Zero-shot transfer to the MRI-only ADNI dataset yields \textit{70.37\% $\pm$ 2.7}, surpassing the best unimodal baseline by \textit{17.2\%}, highlighting robust generalisation under severe missing-modality conditions. Integrated Grad-CAM and SHAP analyses localise saliency to hippocampal and thalamic atrophy and identify eight established AD-related genes, confirming biologically grounded decision-making.

We recognize that our framework is designed to identify statistical associations between input features and Alzheimer’s diagnosis, rather than to determine whether specific biomarkers cause disease progression. Future work could address this limitation by incorporating causal inference methods to better understand underlying mechanisms. Furthermore, our evaluation relies primarily on the ADNI and ANMerge datasets, which may not adequately represent the full spectrum of patient demographics, comorbidities, and clinical environments encountered in real-world settings. This limitation underscores the need to validate our approach on more diverse and representative cohorts. Federated learning offers a promising avenue for future research, as it enables collaborative model training across multiple institutions without requiring direct data sharing, facilitating access to more heterogeneous data sources.

{
    \small
    \bibliographystyle{ieeenat_fullname}
    \bibliography{main}
}

\end{document}